\documentclass{article}

     \PassOptionsToPackage{numbers, compress}{natbib}

\usepackage[preprint]{neurips_data_2024}

\usepackage{graphicx}
\usepackage{subcaption}

\usepackage[utf8]{inputenc} 
\usepackage[T1]{fontenc}    
\usepackage[colorlinks,allcolors=Blue]{hyperref}       
\usepackage{url}            
\usepackage{booktabs}       
\usepackage{amsfonts}       
\usepackage{nicefrac}       
\usepackage{microtype}      
\usepackage[dvipsnames]{xcolor}         

\usepackage{booktabs}
\usepackage{multirow}
\usepackage{cleveref}
\usepackage{enumitem}
\usepackage{listings}
\usepackage{float}
\usepackage{comment}
\usepackage{color}
\usepackage{wrapfig}

\usepackage{ascmac}
\usepackage{framed}

\newcommand{\ie}{\textit{i.e.}}

\newcommand{\comperdial}{\textbf{ComperDial}}
\newcommand{\cpdscore}{\textsc{CPDScore}}

\title{ComperDial: Commonsense Persona-grounded Dialogue Dataset and Benchmark}

\author{
\textbf{Hiromi Wakaki\textsuperscript{$1$}\thanks{~~Equal contribution}\:\:\: Yuki Mitsufuji\textsuperscript{$1 2 \ast$}\:\:\: Yoshinori Maeda$^{1}$\:\:\: Yukiko Nishimura$^{1}$} \\
\textbf{Silin Gao$^{3}$\:\:\: Mengjie Zhao$^{1}$\:\:\: Keiichi Yamada$^{1}$\:\:\: Antoine Bosselut$^{3}$} \\\\
$^1$Sony Group Corporation\:\:\: $^2$Sony AI\:\:\: $^3$EPFL \\
}

\begin{document}

\maketitle
\begin{abstract}

The emergence of large language models as conversational agents has made it critical to reliably assess the open-domain dialogue performance of these systems. However, automatic evaluation of open-domain dialogue remains challenging for multiple reasons: (1) many suitable responses may be appropriate for a given dialogue context, and (2) dialogue response evaluations have traditionally been limited to single-turn responses, while assessing the overall quality of a dialogue requires multiple turns. To address these challenges, we propose a new benchmark, \textbf{Com}monsense \textbf{per}sona-grounded \textbf{Dial}ogue (\textbf{ComperDial}), which facilitates the training and evaluation of \textit{evaluation metrics} for open-domain dialogue systems. 

\comperdial{} consists of human-scored responses for 10,395 dialogue turns in 1,485 conversations collected from 99 dialogue agents submitted to the Commonsense Persona-grounded Dialogue (CPD) challenge. As a result, for any dialogue, our benchmark includes multiple diverse responses with variety of characteristics to ensure more robust evaluation of learned dialogue metrics. In addition to single-turn response scores, \comperdial{} also contains dialogue-level human-annotated scores, enabling joint assessment of multi-turn model responses throughout a dialogue. Finally, building off \comperdial, we devise a new automatic evaluation metric to measure the general similarity of model-generated dialogues to human conversations. Our experimental results demonstrate that our novel metric, \cpdscore{} is more correlated with human judgments than existing metrics. We release both \comperdial{} and \cpdscore{} to the community to accelerate development of automatic evaluation metrics for open-domain dialogue systems.
\end{abstract}

\begin{figure}[t]
    \centering
    \includegraphics[width=1\linewidth]{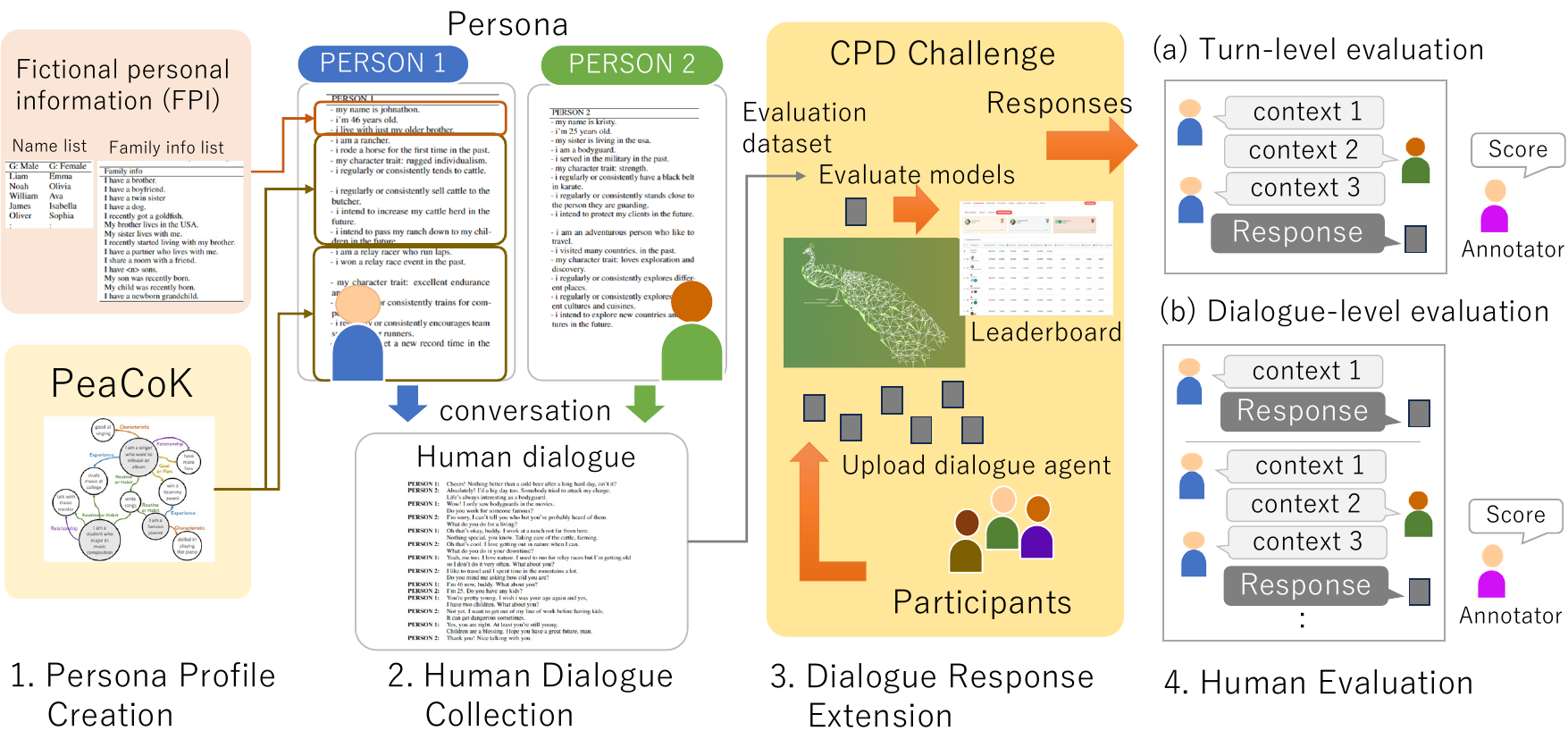}
    \caption{
    \textbf{ Data collection pipeline of \comperdial.} 
   }
    \label{fig:pipeline}
\end{figure}

\section{Introduction}
\label{sec:intro}
Recent advances in large language models (LLMs) have qualitatively improved the open-domain dialogue abilities of conversational agents \citep{openai2023gpt4}, with million users now interacting with chat systems. This adoption has amplified the need for designing agents with dialogue behavior that produces high-quality responses and engages in interesting multi-turn conversations. However, the development of chat agents remains hindered by the challenges of evaluating open-ended dialogues. First, for any given dialogue context, many suitable responses may be appropriate (\ie, one-to-many issue), precluding reliable reference-based evaluation that uses similarity to gold responses as a proxy for quality. Second, dialogue evaluations have traditionally been conducted in single-turn setups where each immediate response is evaluated independent from the full conversation, which misses the important elements of dialogue quality that only emerge over multiple dialogue turns. 

These properties remain challenging to evaluate in a reliable and scalable manner. Automatic evaluation metrics, such as ngram-based metrics ~\cite{papineni-etal-2002-bleu, lin-2004-rouge, banerjee-lavie-2005-meteor, dinan2019second} and embedding-based metrics ~\cite{Zhang2020BERTScore, sellam-etal-2020-bleurt} are highly scalable, but often do not provide faithful assessments of a dialogue's quality. In contrast, human evaluation remains the gold standard of dialogue evaluation (and the baseline to which automatic evaluation metrics are tested for correlation), but is generally slow and expensive, precluding rapid iteration and testing, or based on competitions eliciting human preferences \citep{zheng2023judging,Chiang2024ChatbotAA}, which only assess models against each other without providing assessments of response quality. In response to these challenges, a recent line of work explores developing datasets for training evaluation metrics \cite{mehri-eskenazi-2020-usr, zhao-etal-2020-designing, mehri-eskenazi-2020-unsupervised,zheng2023lmsyschat1m} that assess properties of quality dialogues. Novel LLM-based metrics~\cite{zhang-aaai2024, liu-etal-2023-g, chan2024chateval} evaluated on these datasets have been shown to correlate better with human judgments, but they still fall short of accurately assessing the performance of models.

To address these challenges, we propose a new dataset for training and evaluating open-domain dialogue metrics, \textbf{Com}monsense \textbf{per}sona-grounded \textbf{Dial}ogue dataset (\comperdial), containing human scores for 10,395 dialogue turns and 1,485 full dialogues collected from 97 dialogue systems submitted to the Commonsense Persona-grounded Dialogue (CPD) challenge (along with annotated dialogues from two human subjects). For each dialogue turn, \comperdial{} includes multiple scored responses to capture (and adequately score) the potential diversity of responses in a given context, enabling more robust evaluation of automatic dialogue evaluation metrics. Importantly, every turn is evaluated with respect to the full dialogue context up to that point, provided a holistic assessment of how each response contributes to the dialogue up to that point, allowing non-human response traits (e.g., excessive explanations) and unenjoyable conversation features (e.g., superficial listening) to be evaluated and taken into consideration.
Figure \ref{fig:pipeline} shows the data collection pipeline of \comperdial.

Finally, based on our evaluation scheme, we devise a new automatic evaluation metric,  \cpdscore{} to measure the general likeness of model-generated dialogues to human conversations. \cpdscore{} can be wrapped around any LLM evaluator, and uses Chain-of-Thought~\cite{Wei2022b} reasoning and multi-step prompting to provide auditable explanations of its dialogue assessment. 
Our experimental results show that \cpdscore{} achieves a higher correlation with human scores in \comperdial{} compared to existing dialogue evaluation metrics.

\begin{table}
    \caption{\textbf{Benchmarks for automatic evaluation of dialogue models.} Each column shows a benchmark dataset name and its features. 
    ``No. of dialogue models'' indicates how many dialogue models are used for response generation. ``No. of humans''  indicates how many people are added for response generation to see the performance of human-created responses. 
    }
    \label{tab:benchmarks}
    \centering
    {
    \begin{tabular}{l|ccccc}
    \hline
         & USR~\cite{mehri-eskenazi-2020-usr} &  Zhao et al. ~\cite{zhao-etal-2020-designing} & RADE~\cite{shi2023rade} & FED\cite{mehri-eskenazi-2020-unsupervised} & \comperdial{}\\
    \hline
    \textit{Response collection}    &  &  &  &  & \\
    No. of dialogue models     & 4(3) & 6 & 7 & 2 & \textbf{97}\\
    No. of humans               &  &  &  & 2 & 2\\
    \hline
    \textit{Evaluation technique}     &  &  &  &  & \\
     \ turn-level    & \checkmark & \checkmark & \checkmark & \checkmark & \checkmark\\
     \ dialogue-level    &  &  &  &  & \\
      + interactive eval &  &  &  & \checkmark & \\
      + static dialogue &  &  &  &  & \checkmark \\
    \hline
    \end{tabular}
    }
\end{table}

\section{Related work}
\label{related_work}

\paragraph{Automatic dialogue evaluation} 
For open-domain dialogue research, it is important to use automatic evaluation metrics that ensure efficiency and objective evaluation~\cite{chiang-lee-2023-large}. 
For reference-based metrics, BLEU ~\cite{papineni-etal-2002-bleu}, ROUGE ~\cite{lin-2004-rouge}, 
METEOR ~\cite{banerjee-lavie-2005-meteor}, and F1 ~\cite{dinan2019second} are traditionally used as ngram-based metrics that focus on surface-form similarity. While these metrics provides a simple and general measure, they fail to account for meaning-preserving lexical and compositional diversity. To mitigate this problem, embedding-based metrics such as BERTScore ~\cite{Zhang2020BERTScore} and BLEURT ~\cite{sellam-etal-2020-bleurt} focus on semantic similarity to references without explicit alignment. 
To alleviate the well-known one-to-many issue~\cite{zhao-etal-2017-learning, eval_open_domain_acl} of standard metrics, reference-free metrics with a pre-trained/unsupervised model such as FED~\cite{mehri-eskenazi-2020-unsupervised}, USR~\cite{mehri-eskenazi-2020-usr}, and UniEval ~\cite{zhong-etal-2022-towards} have been proposed. Reference-free metrics that utilize LLMs, such as G-EVAL ~\cite{liu-etal-2023-g}, Zhang et al.~\cite{zhang-aaai2024}, and ChatEval ~\cite{chan2024chateval}, have demonstrated high correlation with human judgement. LLM-based metrics not only require no reference but also have the advantage of defining multiple aspects similar to human evaluation. 
In contrast to the existing work, our \cpdscore{} can incorporate a comprehensive consideration of multiple aspects when rating a single overall score.

\paragraph{Human evaluation} 
As the conventional automatic evaluation metrics are not fully reliable in terms of the one-to-many issue, both automatic evaluation and human evaluation are usually conducted ~\cite{zhang-etal-2018-personalizing, gopalakrishnan2019topical, chen2023learning}. 
To check the fine-grained quality of models with multiple aspects, fluency~\cite{Liu_Huang_Zhang_Wang_deMelo_Lin_Pang_He_2023, shi2023rade, macina2023opportunities, zhang-etal-2018-personalizing, gao-etal-2023-peacok}, engagingness~\cite{shi2023rade, zhang-etal-2018-personalizing, zhang-aaai2024, gao-etal-2023-peacok, shuster2022blenderbot}, consistency~\cite{Liu_Huang_Zhang_Wang_deMelo_Lin_Pang_He_2023, zhang-etal-2018-personalizing, gao-etal-2023-peacok, shuster2022blenderbot}, coherence~\cite{Liu_Huang_Zhang_Wang_deMelo_Lin_Pang_He_2023, macina2023opportunities, zhang-aaai2024}, informativeness~\cite{zhang-aaai2024, thoppilan2022lamda}, and relevance~\cite{shi2023rade, zhang-aaai2024} are commonly utilized as aspects. 
In \comperdial, we annotate a single overall score as human evaluation based on six major aspects that relate to human-like response quality: fluency, consistency, coherency, engagingness, persona consistency, and humanness.

\paragraph{Evaluation techniques}
To consider multi-turn diversity and quality, interactive evaluation is conducted as dialogue-level evaluation~\cite{svikhnushina-etal-2022-ieval, see-etal-2019-makes, relaible-2022, mehri-etal-2022-interactive, mehri-eskenazi-2020-unsupervised, smith-etal-2022-human}. Human workers are asked to chat with models to collect conversations between humans and models via their own systems. 
Chatbot Arena~\cite{zheng2023judging}, which is not designed for open-domain dialogue but rather for LLM quality assessment, is a crowdsourced platform featuring anonymous battles between LLMs.
Thus, human-bot interactive evaluation requires human-in-the-loop, which can be a costly process. 
Another problem is that the comparison conditions cannot be the same between models because each conversation is dependent on the workers. 
Even if interactive evaluation is automated through self-play ~\cite{ghandeharioun2019approximating} and bot-bot~\cite{chatmatch}, it is still challenging to compare models fairly under the same conditions because the dialogue history will differ between evaluators.
In contrast, \comperdial{}'s dialogue-level annotation considers multi-turn diversity and quality under the same conditions between models instead of interactive evaluation. 

\paragraph{Dialogue evaluation benchmarks}
USR~\cite{mehri-eskenazi-2020-usr} is commonly used as benchmark dataset for automatic dialogue evaluation. 
While USR-Persona collects responses using three models on PersonaChat~\cite{zhang-etal-2018-personalizing}, USR-Tocical collects generated responses using one model with four different conditions on TopicalChat~\cite{gopalakrishnan2019topical}. Zhao et al. ~\cite{zhao-etal-2020-designing} use six generative models on PersonaChat and DailyDialog~\cite{li-etal-2017-dailydialog}. RADE ~\cite{shi2023rade} collects responses using seven generative models on DSTC-ChitChat~\cite{hori2018endtoend}, Empathetic Dialogue~\cite{rashkin-etal-2019-towards}, and PersonaChat. In contrast, FED\cite{mehri-eskenazi-2020-unsupervised} uses two generative models to create original dialogues between a human and a human and between a human and two models to enable interactive evaluation. Our \comperdial{} contains responses generated by 97 conversational agents and two human workers so that we can thoroughly analyze the collation between human judgement and automatic evaluation as system-level scores.

\begin{table}[h]
    \caption{ \textbf{Example dialogue from the \comperdial{} dataset.} We provide the full dialogue in Appendix \ref{sec:app_comperdial}.}
    \label{tab:dialogue_example}
    \centering
    {\small
    \begin{tabular}{p{30mm} p{30mm}}
    \hline
      PERSON 1  &  PERSON 2\\
    \hline
    - my name is johnathon.	&	- my name is kristy.	\\
    - i'm 46 years old.	&	- i'm 25 years old.	\\
    - i live with just my older brother.	&	- my sister is living in the usa.	\\
    - i am a rancher.	&	- i am a bodyguard.	\\
    - i rode a horse for the first time in the past.	&	- i served in the military in the past.	\\
     ...   &  ... \\ 
    \hline
    \end{tabular}
    \begin{tabular}{p{3mm} p{57mm}}
    \textbf{P1} & Cheers! Nothing better than a cold beer after a long hard day, isn't it? \\
\textbf{P2} & Absolutely! I'd a big day too. Somebody tried to attack my charge. Life's always interesting as a bodyguard.\\
\textbf{P1} & Wow! I only saw bodyguards in the movies. Do you work for someone famous? \\
\textbf{P2} & I'm sorry, I can't tell you who but you've probably heard of them. What do you do for a living?\\
\textbf{P1} & Oh that's okay, buddy. I work at a ranch not far from here. Nothing special, you know. Taking care of the cattle, farming.  \\
    \end{tabular}
    }
\end{table}

\section{\comperdial}
\label{sec:comperdial}
\comperdial{} contains a new dialogue dataset assigned with persona sentences in a similar format to PersonaChat (Table \ref{tab:dialogue_example}). Specifically, to ensure that the interlocutors' utterances are consistent throughout each collected dialogue, we create a persona profile for each interlocutor following the idea of PersonaChat. 
Our benchmark creation process consists of two main steps: persona profile creation 
(Section \ref{sec:persona_profile}) and diversified dialogue collection (Section \ref{sec:dialogue_collection}). In \ref{sec:dialogue_collection}, we begin by collecting dialogues from humans. These human dialogues are then used as context to prompt more diversified dialogues from various chat models as an expansion of multiple possible responses. Finally, a human evaluation is conducted to check the quality of the model-generated dialogues/responses.

\subsection{Persona profile creation}
\label{sec:persona_profile}
We create the main part of the persona profiles based on PeaCoK~\cite{gao-etal-2023-peacok}, where head personas and their tail attributes are extracted from PeaCoK and converted to natural language sentences as statements/items in the persona profiles.  
We first select a head persona, and then obtain a tail persona from each of eight aspects, namely, Characteristic, RoutineHabit, GoalPlan, and Experience with/without Relationship.\footnote{Because there are not necessarily eight types, we used only aspects where there was at least one tail persona.} 
We process on PeaCoK head/tail entities to ensure a better quality of created profiles as follows: (a) When selecting a head persona, in the case of a negative impression, we skip it (e.g. ``forger'', ``dishonest person'')\footnote{When asking workers to play the role of a type of person that people generally feel negative about, it's better to ask them whether or not they are okay with including that condition.} (b) When obtaining a tail persona, in the case of contradictory sentences, correct them to make them consistent. (c) In the case of gendered expressions, change them to gender-neutral expressions. (e.g., ``police man'' -> ``police officer'')\footnote{Some words cannot be changed because there are no alternative expressions (e.g., king, queen)}

Since PeaCoK does not contain concrete personal information such as name, age, etc., we add new sentences to define fictional personal information (FPI) for natural conversation creation.
The format to create fictional persona information is as follows:``\textit{My name is X. I'm Y years old. R\lbrack Family Info\rbrack.}''.
Details on how to create the FPI are provided in Appendix \ref{sec:app_comperdial}.

We create one persona by combining one FPI and two persona profiles. These items are randomly selected and combined as a list of persona sentences. If there are contradictory sentences as a result, we correct them to make them consistent. For example, if there is a sentence such as ``\textit{I worked for a company for 20 years}'' in the persona profile and ``\textit{I'm 25 years old}'' in the FPI, these two sentences are contradictory and we change the FPI to one that is better matched to the situation.
Overall, we make sure that our procedures yield persona profiles that fit real-life scenarios and are commonsense.

\subsection{Diversified dialogue collection}
\label{sec:dialogue_collection}
\subsubsection{Human dialogue collection}
Based on the personas, we collect new dialogues in the form of role play conversations carried out by human workers. 
This human dialogue collection is performed with 40 workers who passed a qualified check.
Details on the qualified check are provided in Appendix \ref{sec:app_comperdial}.
As it is difficult to ensure the quality of crowdsourced workers, who are often anonymous participants or volunteers, we have chosen not to utilize a crowdsourcing platform in this study.
The data creation is conducted in a similar style to an online chat. Each interlocutor knows their own persona setting, but they are unaware of the corresponding interlocutor's persona setting. The purpose of the conversation is to get to know each other.  We encourage them to imagine each situation on their own. In cases where a person finds it difficult to imagine a situation where they meet an unknown person and engage in conversation, the workers are asked to imagine having a conversation with an avatar or a character in a virtual space or game.

\subsubsection{Diversified dialogue response extension}
\label{sec:response_extension}
We perform the dialogue response expansion by collecting model responses from Task 1 of the CPD challenge.
The competition aims to identify the best approach among state-of-the-art participant dialogue models using our human dialogue collection as an evaluation dataset.
Based on the assigned personas of two interlocutors, the task is to develop a dialogue model that generates one interlocutor’s response to their counterpart, given the dialogue history between these two interlocutors. 

There are two tracks: the GPU track and the prompt engineering track. The GPU track aims to run participants' LLMs on Amazon AWS\footnote{The node is g5.2xlarge.}.The prompt engineering track aims to use OpenAI GPT-3.5 with the participants’ own prompt engineering.\footnote{The API version is gpt-3.5-turbo-0125.}

\subsubsection{Human evaluation}
\label{sec:human_evaluation}
Annotation of the human evaluation is performed by six internal workers who are not researchers but assistants to researchers. All are English speakers. They have received training in dialogue evaluation by observing various dialogue models and learning different aspects of dialogue evaluation. To ensure a high inter-agreement rate among the annotators, crowdsourcing is not utilized. 

Annotators are asked to assign an overall score from 1 to 5.
In the \comperdial{} dataset, the static single-turn evaluation and the proposed static multi-turn / dialogue-level evaluation are conducted using a different evaluation process to determine whther the way the dialogue is presented affects the evaluation. See et al. ~\cite{see-etal-2019-makes} concluded that a chatbot does not need to be human-like to be enjoyable because, while their models failed to get close to actual humans in terms of humanness, they achieved close-to-human scores on engagingness. However, just as visual reality is being pursued in images, there is also a need for conversational agents to align with human reality. Therefore, we define a kind of humanness as the overall score of evaluation. Note, however, that the definition and expectation of high scores for humanness or human-like responses can vary from person to person. Additionally, the quality of dialogue has multiple aspects, so when evaluating dialogue models, it is important to consider more than one quality metric ~\cite{see-etal-2019-makes}. Thus, we also take into account specific aspects when evaluating the overall score. For the explanation of the overall score, we include all six aspects of fluency, consistency, coherence, engagingness, persona consistency, and humanness as indicators of high quality.

\begin{table}[h]
    \caption{\textbf{Correlation scores on \comperdial{} for turn-level evaluation.} (1) Scores obtained from a single API call using the original prompts and (2) scores from a single API call using \cpdscore{} prompts are shown. We provide the results using GPT-3.5 in Appendix \ref{sec:app_experiment}. }
    \label{tab:original_once}
    \centering
    {\small
    \begin{tabular}{lcccc}
        \hline
        \multirow{2}{*}{\textbf{Methods}} & \multicolumn{2}{c}{Turn-level  $\uparrow$} &   \multicolumn{2}{c}{System-level  $\uparrow$}  \\
         \cline{2-3} \cline{4-5} 
         & $\rho$ & $\tau$ &   $\rho$ & $\tau$  \\
                \hline
                \colorbox{gray}{\textcolor{white}{Original + once }} &  &  &  &   \\
                \ \textbf{Original (GPT-4)} &  &  &  &     \\
                \ Simple &  0.547 & 0.454 &   0.820  & 0.652 \\
                \ Detail &  0.237 & 0.191 &   0.674  & 0.535 \\
                \hline
                \colorbox{gray}{\textcolor{white}{\cpdscore{} + once }} &  &  &  &   \\
                \ \textbf{Ours (GPT-4)} &  &  &  &     \\
                \ CPDS-S w/ ref  &  0.597 & 0.490 &   0.884  & 0.724 \\
                \ CPDS-S w/o ref  &  0.615 & 0.517 &   0.896  & 0.746 \\
                \ CPDS-D w/ ref  &  0.629 & 0.523 &   0.873  & 0.710 \\
                \ CPDS-D w/o ref &  \textbf{0.681} & \textbf{0.576}  & \textbf{0.923}  & \textbf{0.790} \\
                \hline
              
    \end{tabular}
    }
\end{table}

\section{\cpdscore{}}
\label{sec:cpdscore}
We utilize a new LLM-based evaluation metric called \textbf{\cpdscore{}}, which is based on the same criteria as the annotation criteria.
Our prompts, similar to the guidelines of human annotation, focus on an overall score based on humanness taking into account several aspects described in Section \ref{sec:human_evaluation}. 
For \textit{turn-level} evaluation, we define two types of description: a simple prompt and a detailed prompt. The simple prompt is a variant of Zhang et al.'s~\cite{zhang-aaai2024}, and the detailed prompt is an variant of G-EVAL~\cite{liu-etal-2023-g}. We call these turn-level \textbf{\cpdscore{}-Simple} and \textbf{\cpdscore{}-Detail}, respectively. 
We also define two types of prompts, one with and one without a gold response as a reference response, and check whether it is effective to include the gold response as a reference in the prompt.
For \textit{dialogue-level} evaluation, we newly design a description to consider multiple responses in a dialogue. We refer to it as \textbf{\cpdscore{}-Dialogue}.
\cpdscore{}-Dialogue is a two-step evaluation method. In step 1, each turn is evaluated for turn-level performance using \cpdscore{}-Detail/Simple. In Step 2, first, an intermediate result is generated by assessing the all-turn-results in a dialogue from Step 1. Next, the responses generated for each turn are collectively checked with multi-turn evaluation criteria such as non-human-like traits and superficial responses. Finally, based on this evaluation, adjustments are made by adding or subtracting points to the previous intermediate result for further refinement.
We provide the details of the prompts in Appendix \ref{sec:app_promt}. 
We also compare OpenAI GPT-4\footnote{The API version is gpt-4-turbo-2024-04-09.} with GPT-3.5\footnote{The API version is gpt-3.5-turbo-0125.} using the same prompts. If nothing is specified, we use the average of the results from three API calls.

\section{Benchmark data analysis}

\paragraph{Response collection}
We used 250 dialogues from the newly created dataset for evaluation in Round 1 of CPD Challenge Task 1. However, considering the annotation cost, we selected just 15 dialogues out of the 250. 
In addition to model-generated responses, we also included human-created responses from two internal workers to compute the scores of human created responses. These two workers are native English speakers and different from the annotators mentioned in Section \ref{sec:human_evaluation}, as the annotators should not know which responses were created by humans. Thus, \comperdial{} contains 15 dialogues from Round 1 including responses generated by 97 conversational agents and the two human workers. All dialogues comprise seven turns each. The total number of responses was 10,395.
This provides a sufficient amount of data for turn-level scores, dialogue-level scores, and system-level scores: 10,395 turns, 1,485 dialogues, and 99 systems, respectively.

\paragraph{Inter-annotator agreement}
To determine the inter-annotator agreement of annotation in \comperdial{}, we computed Krippendorff’s Alpha~\cite{krippendorff2018content} for turn-level evaluation and dialogue-level evaluation.
Krippendorff’s Alpha was 0.56 for turn-level evaluation and 0.62 for dialogue-level evaluation. 
These results indicate a high level of inter-annotator agreement, with values exceeding 0.4.

\section{Experimental results}
\label{sec:experimental_results}

\subsection{Baseline metrics}
\label{sec:baseline}
We evaluate the effectiveness of different automatic metrics for dialogue models on \comperdial{} and other benchmark datasets to compare our \cpdscore{} with existing baseline metrics.
For ngram-based metrics, \textbf{BLEU} ~\cite{papineni-etal-2002-bleu}, \textbf{ROUGE} ~\cite{lin-2004-rouge}, 
\textbf{METEOR} ~\cite{banerjee-lavie-2005-meteor}, and \textbf{FI} ~\cite{dinan2019second} are evaluated. These metrics consider surface-form similarity with a gold utterance. For embedding-based metrics, \textbf{BERTScore} ~\cite{Zhang2020BERTScore} and \textbf{BLEURT} ~\cite{sellam-etal-2020-bleurt} are evaluated. Embedding-based metrics consider the semantic-level similarity with a gold utterance.
For pretrained model based metrics, \textbf{FED}~\cite{mehri-eskenazi-2020-unsupervised} and \textbf{UniEval} ~\cite{zhong-etal-2022-towards} are evaluated.

\begin{table}[h]
    \caption{\textbf{Correlation scores on \comperdial{} for turn-level evaluation.} The correlation scores of the Pearson's correlation ($r$), Spearman's correlation ($\rho$), and Kendall's correlation ($\tau$) between turn-level ranking and system-level ranking on turn-level human evaluation and automatic evaluation are shown. 
    All values are statistically significant to p-value < 0.05 unless marked by $^{*}$. The FED$_{En}$ and FED$_{Rel}$ indicate two evaluation perspective of FED, i.e., engagement and relevance. 
    \textbf{Bold} the best results of all methods when using GPT-4 as the \cpdscore{}, and \underline{underline} the best results of all methods when using GPT-3.5 as the \cpdscore{}.
     }

    \label{tab:bench_result_turn}
    \centering
    {\small
    \begin{tabular}{lcccccc}
        \hline
        \multirow{2}{*}{\textbf{Methods}} & \multicolumn{3}{c}{Turn-level score $\uparrow$} &   \multicolumn{3}{c}{System-level score $\uparrow$}  \\
         \cline{2-4} \cline{5-7} 
         & $r$ & $\rho$ & $\tau$ &  $r$ & $\rho$ & $\tau$  \\

                \hline
                \colorbox{gray}{\textcolor{white}{Based on turn-level evaluation}} &  &  &  &  &  &   \\
                \ BLEU~\cite{papineni-etal-2002-bleu} & 0.104& 0.159 & 0.113 &  0.392 & 0.206  & 0.136 \\
                \ F1 ~\cite{dinan2019second} & 0.196& 0.185 & 0.132 &  0.584 & 0.120$^{*}$  & 0.078$^{*}$ \\
                \ METEOR~\cite{banerjee-lavie-2005-meteor} & 0.197& 0.201 & 0.143 &  0.766 & 0.556  & 0.410 \\
                \ ROUGE~\cite{lin-2004-rouge} & 0.187& 0.178 & 0.127 &  0.526 & 0.057$^{*}$  & 0.028$^{*}$ \\
                \ BERTScore~\cite{Zhang2020BERTScore} & 0.269& 0.227 & 0.159 &  0.369 & 0.075$^{*}$  & 0.044$^{*}$ \\
                \ BLEURT~\cite{sellam-etal-2020-bleurt} & 0.347& 0.331 & 0.236 &  0.810 & 0.873  & 0.699 \\
                \ FED$_{En}$~\cite{mehri-eskenazi-2020-unsupervised} & -0.009$^{*}$& 0.043 & 0.030 &  0.569 & 0.585  & 0.470 \\
                \ FED$_{Rel}$~\cite{mehri-eskenazi-2020-unsupervised} & -0.025& -0.025 & -0.017 &  0.399 & 0.378  & 0.247 \\
                \ UniEval~\cite{zhong-etal-2022-towards} & 0.397& 0.404 & 0.287 &  0.637 & 0.584  & 0.432 \\
                \hline
                \ CPDS-S w/ ref (GPT-3.5) & 0.507& 0.489 & 0.389 &  0.773 & 0.879  & 0.717 \\
                \ CPDS-S w/ ref (GPT-4) & 0.678& 0.629 & 0.493 &  \textbf{0.946} & 0.900  & 0.742 \\
                \ CPDS-S w/o ref (GPT-3.5) & 0.491& 0.468 & 0.371 &  0.715 & 0.861  & 0.684 \\
                \ CPDS-S w/o ref (GPT-4) & 0.688& 0.662 & 0.532 &  0.924 & 0.899  & 0.753 \\
                \ CPDS-D w/ ref (GPT-3.5) & \underline{0.596}& \underline{0.575} & \underline{0.445} &  \underline{0.890} & \underline{0.890}  & \underline{0.740} \\
                \ CPDS-D w/ ref (GPT-4) & 0.689& 0.656 & 0.524 &  0.939 & 0.886  & 0.727 \\
                \ CPDS-D w/o ref (GPT-3.5) & 0.592& 0.564 & 0.435 &  0.886 & 0.876  & 0.717 \\
                \ CPDS-D w/o ref (GPT-4) & \textbf{0.714}& \textbf{0.712} & \textbf{0.583} &  0.943 & \textbf{0.928}  & \textbf{0.799} \\
                \hline

    \end{tabular}
    }
\end{table}

\begin{table}
    \caption{\textbf{Correlation scores on \comperdial{} for dialogue-level evaluation.} Notations are the same as Table \ref{tab:bench_result_turn}.``Human'' indicates the results when using annotations for turn-level evaluation, which are the gold labels in Table \ref{tab:bench_result_turn}. 
    }
    \label{tab:bench_result_dialogue}
    \centering
    {\small
    \begin{tabular}{lcccccc}
        \hline
        \multirow{2}{*}{\textbf{Methods}} & \multicolumn{3}{c}{Dialogue-level score $\uparrow$} &   \multicolumn{3}{c}{System-level score $\uparrow$}  \\
         \cline{2-4} \cline{5-7} 
         & $r$ & $\rho$ & $\tau$ &  $r$ & $\rho$ & $\tau$  \\

                \hline
                \colorbox{gray}{\textcolor{white}{Based on turn-level evaluation}} &  &  &  &  &  &   \\
                \ BLEU~\cite{papineni-etal-2002-bleu} & 0.122& 0.181 & 0.127 &  0.363 & 0.191$^{*}$  & 0.120$^{*}$ \\
                \ F1 ~\cite{dinan2019second} & 0.194& 0.114 & 0.079 &  0.476 & 0.138$^{*}$  & 0.093$^{*}$ \\
                \ METEOR~\cite{banerjee-lavie-2005-meteor} & 0.293& 0.231 & 0.163 &  0.674 & 0.535  & 0.397 \\
                \ ROUGE~\cite{lin-2004-rouge} & 0.165& 0.090 & 0.061 &  0.414 & 0.071$^{*}$  & 0.039$^{*}$ \\
                \ BERTScore~\cite{Zhang2020BERTScore} & 0.219& 0.113 & 0.077 &  0.278 & 0.092$^{*}$  & 0.059$^{*}$ \\
                \ BLEURT~\cite{sellam-etal-2020-bleurt} & 0.448& 0.354 & 0.253 &  0.697 & 0.826  & 0.647 \\
                \ FED$_{En}$~\cite{mehri-eskenazi-2020-unsupervised} & 0.086& 0.049$^{*}$ & 0.035$^{*}$ &  0.484 & 0.550  & 0.434 \\
                \ FED$_{Rel}$~\cite{mehri-eskenazi-2020-unsupervised} & 0.108& 0.125 & 0.088 &  0.444 & 0.359  & 0.232 \\
                \ UniEval~\cite{zhong-etal-2022-towards} & 0.418& 0.402 & 0.290 &  0.503 & 0.548  & 0.388 \\
                \hline
                \ CPDS-S w/ ref (GPT-3.5) & 0.592& 0.636 & 0.486 &  0.693 & 0.861  & 0.685 \\
                \ CPDS-S w/ ref (GPT-4) & 0.782& 0.751 & 0.589 &  0.902 & 0.894  & 0.728 \\
                \ CPDS-S w/o ref (GPT-3.5) & 0.513& 0.525 & 0.391 &  0.617 & 0.850  & 0.666 \\
                \ CPDS-S w/o ref (GPT-4) & 0.743& 0.718 & 0.561 &  0.871 & 0.902  & 0.750 \\
                \ CPDS-D w/ ref (GPT-3.5) & 0.671& \underline{0.670} & \underline{0.513} &  0.783 & 0.867  & 0.705 \\
                \ CPDS-D w/ ref (GPT-4)  & \textbf{0.792}& 0.761 & 0.601 &  0.905 & 0.889  & 0.727 \\
                \ CPDS-D w/o ref (GPT-3.5) & 0.657& 0.646 & 0.490 &  0.781 & 0.856  & 0.689 \\
                \ CPDS-D w/o ref (GPT-4) & 0.776& \textbf{0.768} & \textbf{0.610} &  0.899 & \textbf{0.926}  & \textbf{0.784} \\
                \hline
                \ Human & 0.855& 0.827 & 0.676 &  0.963 & 0.968  & 0.869 \\
                \hline
                \colorbox{gray}{\textcolor{white}{Based on dialogue-level evaluation}} & &  &  &  &  &  \\   
                \ CPDS-Dial (GPT-3.5)  & \underline{0.675} & 0.656 & 0.501 & \underline{0.818} & \underline{0.873} & \underline{0.715}\\
                \ Intermediate score  & 0.664 & 0.658 & 0.506 & 0.779 & 0.868 & 0.710\\
                \ CPDS-Dial (GPT-4) & 0.791 & 0.742 & 0.587 & \textbf{0.926} & 0.924 & 0.783   \\     
                \ Intermediate score   & 0.777 & 0.763 & 0.607 & 0.903 & 0.925 & 0.781\\
                \hline
    \end{tabular}
    }
\end{table}

\subsection{Metric correlation under turn-level evaluation}
Table \ref{tab:bench_result_turn} shows the benchmark results on \comperdial{} for \textit{turn-level} evaluation, where a turn-level score is assigned to each turn.
These results include turn-level scores and system-level scores.\footnote{A system level score is calculated by averaging turn-level scores of all turns in a system.} 
In both turn-level and system-level scores, the majority of the results showed high reliability with p-values < 0.05. Among the baseline metrics, BLEURT showed the best correlation. All \cpdscore{} except CPDS-S without a reference response using GPT-3.5 outperformed all baseline metrics.

Next, when comparing different conditions using \cpdscore{}, we found that GPT-4 achieved higher scores than GPT-3.5. The correlation results of CPDS-D were also higher than those of CPDS-S. In contrast, when comparing models with and without a reference response, except for the case of using GPT-4 with detail prompts, the models with a reference response generally performed better. This suggests that CPDS-D with GPT-4, which has a strong ability to understand context, can achieve more human-like evaluations even without a reference response. The best case was CPDS-D without a reference response using GPT-4, and its Spearman's correlation and Kendall’s correlation were 0.928 and 0.799 respectively. Considering the current API usage cost, where GPT-4 is more expensive than GPT-3.5, the cost-effective approach would be to use CPDS-D with a reference response using GPT-3.5, which still achieves a high correlation (0.890 and 0.740).

Table \ref{tab:original_once} presents the results of the correlation coefficients, based on (1) scores obtained from a single API call using the original prompts and (2) scores from a single API call using \cpdscore{} prompts.\footnote{The results of \cpdscore{} in Tables \ref{tab:bench_result_turn} and \ref{tab:bench_result_dialogue} are based on the averages scores of three API calls.} 
When comparing the results of \cpdscore{} between the number of AIP calls, we observed that all cases averaging three API call results outperformed those of a single API call. CPDS-D using GPT-4 without a reference response also achieved the highest performance in the case of a single API call, and even a single AIP call was sufficient to yield high scores without three-time AIP calls. For the original prompt of CPDS-S, which originally calculated the overall score, the correlation was 0.821 using GPT-4, and for the original prompt of CPDS-D, which originally calculated engagingness, the correlation was 0.642 using GPT-4. 
\emph{All of the cases of \cpdscore{} showed higher correlations with human judgement than the original prompts.}

\subsection{Metric correlation under dialogue-level evaluation}
Table \ref{tab:bench_result_dialogue} shows the results on \comperdial{} for \textit{dialogue-level} evaluation where a dialogue-level score is assigned to each dialogue. 
In the case of turn-level evaluation based metrics, the dialogue-level scores are calculated by averaging the turn-level scores of all turns in a dialogue. 
This table shows both dialogue-level scores and system-level scores.\footnote{The system level score is calculated by averaging turn-level (or dialogue) scores of all turns (or dialogues) in a system.} 
For metrics based on turn-level evaluation, comparing the correlations in the system-level score of each method of Table \ref{tab:bench_result_dialogue} with those of Table \ref{tab:bench_result_turn} showed an overall decrease in the correlation of Table \ref{tab:bench_result_dialogue}.
These results suggest that, since these metrics are based on turn-level evaluation, the correlation observed in Table \ref{tab:bench_result_dialogue}, which was annotated based on dialogue-level evaluation, was lower compared to Table \ref{tab:bench_result_turn}, which was annotated based on turn-level evaluation.
For metrics based on dialogue-level evaluation, CPDS-Dial (GPT-3.5) used the all-turn-results of CPDS-D w/ ref (GPT-3.5) as Step 1, and CPDS-Dial (GPT-4) used those of CPDS-D w/o ref (GPT-4) as Step 1. 
When comparing the results using GPT-3.5, the results of validating CPDS-Dial reveal that, while the correlation in the dialogue-level score is lower compared to the turn-level-based metrics, CPDS-Dial exhibits the higher correlation in the system-level scores than tern-level evaluation based CPDS-S/D. Similarly, when comparing CPDS-Dial, which employs a two-step approach, with its intermediate score, it is confirmed that the two-step approach contributes to improvements in the system-level scores since the correlation results of CPDS-Dial outperformed its intermediate score in the system-level scores. 
In contrast, when comparing the results using GPT-4, CPDS-D w/o ref (GPT-4) showed the best Spearman’s and Kendall’s correlation on both dialogue-level and system-level scores. CPDS-Dial could improve only Pearson's correlation from its intermediate score and CPDS-D w/o ref (GPT-4) that is used in the Step 1.
While the results of these experiments using GPT-3.5 suggest that the two-step approach of CPDS-Dial can improve the correlation between human judgments in dialogue-level evaluations when system-level scores are checked, those using GPT-4 showed there was little change in performance due to the different approaches.

\begin{table}[h]
    \caption{\textbf{Correlation scores on USR-TopicalChat and USR-PersonaChat ~\cite{mehri-eskenazi-2020-usr} for turn-level evaluation.} Metrics $r$ and $\rho$ indicate Pearson's $r$ and Speaman's $\rho$. We provide the results using GPT-3.5 in Appendix \ref{sec:app_experiment}.}
    \label{tab:usr_result}
    \centering
    {\small
    \begin{tabular}{lccccc}
        \hline
        \multirow{2}{*}{\textbf{Methods}} & \multicolumn{2}{c}{USR-Topical} & & \multicolumn{2}{c}{USR-Persona} \\
        \cline{2-3} \cline{5-6}
         & $r$ & $\rho$ && $r$ & $\rho$\\
        \hline
        \textbf{Baseline ~\cite{shi2023rade}} &  &  &  &  & \\
        \ METEOR & 0.336 & 0.391 &  & 0.253 & 0.271\\
        \ BERTScore & 0.298 & 0.325 &  & 0.152 & 0.122\\
        \ BLEURT & 0.216 & 0.261 &  & 0.065 & 0.054\\
        \ RADE & 0.480 & 0.466 &  & 0.451  & 0.465\\
        \ GRADE & 0.200 & 0.217 &  & 0.358 & 0.352\\
        \ USR & 0.412 & 0.423 & & 0.440  & 0.418 \\
        \ USL-H & 0.322 & 0.340 &  & 0.495 & 0.523\\
        \hline
                \ \textbf{Ours (GPT-4)} &  &  &  &  &    \\
                \ CPDS-S w/ ref  & 0.668& 0.652 &  & 0.631 & 0.628 \\
                \ CPDS-S w/o ref & 0.663& 0.661 &  & \textbf{0.693} & \textbf{0.681} \\
                \ CPDS-D w/ ref  & 0.664& 0.649 &  & 0.634 & 0.638 \\
                \ CPDS-D w/o ref  & \textbf{0.681}& \textbf{0.667} &  & 0.646 & 0.645 \\
                \hline

    \end{tabular}
    }
\end{table}

\subsection{Results on USR datasets}
\label{sec:usr}
We further evaluated the robustness of \cpdscore{} by using the existing dialogue evaluation benchmarks, namely two USR datasets ~\cite{mehri-eskenazi-2020-usr}. The results are shown in Table \ref{tab:usr_result}, where we can see that \cpdscore{} outperformed all other metrics.
Consistent with the \comperdial{} results, GPT-4 outperformed GPT-3.5. However, for both USR-Topical and USR-Persona datasets, under the same conditions, the performance was consistently higher when not using a reference response compared to when using one.
This finding indicates that while reference responses are generally advantageous in \comperdial{}, their utility may differ across dialogue datasets.
In terms of prompt types, CPDS-D, as in \comperdial{}, showed a superior overall performance compared to CPDS-S.

\section{Conclusion}
\label{sec:conclusion}
In this paper, we proposed \comperdial{} as a benchmark of automatic dialogue evaluation metrics to cover diverse responses and a sufficient number of systems for reliability. \comperdial{} contains 99 systems including 97 conversational agents and two human workers. Our human evaluation annotation consists of a static single-turn evaluation and a static multi-turn/dialogue level evaluation. We assessed eight existing baseline metrics and \cpdscore{} on \comperdial{}, and the findings showed that \cpdscore{} had a high correlation with human judgement on the benchmark. 
A limitation of this work is the condition of the metric on the leaderboard of the CPD Challenge, where we collected responses generated by models. Since we used Word F1 on the leaderboard of Round 1, the models submitted in Round 1 might have been aimed at getting a higher score on Word F1. In future work, we plan to create an additional benchmark dataset of Round 2 where the metrics have been changed, thereby helping people develop more robust metrics.

\begin{ack}
The data collection was conducted on the AIcrowd site. This work was carried out with the support of 
Sharada Mohanty, Dipam Chakraborty, and Sneha Nanavati at AIcrowd. We also wish to thank all participants of the CPD challenge.
\end{ack}

\newpage
\bibliographystyle{unsrt}
\bibliography{reference}

\newpage
\appendix
\tableofcontents
\newpage
\section{ComperDial}
\label{sec:app_comperdial}

\subsection{URL}
\href{https://huggingface.co/datasets/Sony/ComperDial}{https://huggingface.co/datasets/Sony/ComperDial}

\subsection{Statistics of benchmarks}

To compare automatic evaluation methods such as ngram-based metrics ~\cite{papineni-etal-2002-bleu, lin-2004-rouge, banerjee-lavie-2005-meteor, dinan2019second}, embedding-based metrics ~\cite{Zhang2020BERTScore, sellam-etal-2020-bleurt}, and LLM-based metrics ~\cite{zhang-aaai2024, liu-etal-2023-g, chan2024chateval}, existing benchmarks \cite{mehri-eskenazi-2020-usr, zhao-etal-2020-designing, mehri-eskenazi-2020-unsupervised} for the automatic evaluation of dialogue models are typically used (Tables \ref{tab:benchmarks_2} and \ref{tab:number_of_examples_bench}). 

\begin{table}[h]
    \caption{\textbf{Benchmarks for automatic evaluation of dialogue models.} Each column shows a benchmark dataset name and its features. 
    "\# of dialogue models" indicates how many dialogue models are used for response generation. "\# of human" in that indicates how many people are added for response generation to see the performance of human created responses. }
    \label{tab:benchmarks_2}
    \centering
    \begin{tabular}{l|ccccc}
    \hline
         & USR~\cite{mehri-eskenazi-2020-usr} & Zhao~\cite{zhao-etal-2020-designing}  & RADE~\cite{shi2023rade} & FED\cite{mehri-eskenazi-2020-unsupervised} & ComperDial\\
    \hline
    \textit{Dialogue dataset}     &  &  &  &  & \\
     \ Persona Chat~\cite{zhang-etal-2018-personalizing}     & \checkmark & \checkmark & \checkmark  &  & \\
     \ Topical Chat~\cite{gopalakrishnan2019topical}     & \checkmark &  &  &  & \\
     \ Daily Dialogue~\cite{li-etal-2017-dailydialog}    &  & \checkmark &  &  & \\
     \ Empathetic Dialogue~\cite{rashkin-etal-2019-towards}   &  &  & \checkmark &  & \\
     \ DSTC-ChitChat~\cite{hori2018endtoend}    &  &  & \checkmark &  & \\
     \ \textbf{Newly Created Dialogue}    &  &  &  & \checkmark & \checkmark \\
    \hline
    \textit{Response collection}    &  &  &  &  & \\
    \# of dialogue models     & 4(3) & 6 & 7 & 2 & \textbf{97}\\
    \# of human               &  &  &  & 2 & 2\\
    \hline
    \textit{Evaluation technique}     &  &  &  &  & \\
     \ turn-level    & \checkmark & \checkmark & \checkmark & \checkmark & \checkmark\\
     \ dialouge-level    &  &  &  &  & \\
      + interactive eval &  &  &  & \checkmark & \\
      + static dialogue &  &  &  &  & \checkmark \\
    \hline
    \end{tabular}

\end{table}

\begin{table}[h]
    \caption{The number of examples when calculating turn-level, dialogue-level, and system-level score correlations. "N/A" indicates that the data set does not provide the required information to calculate the level score. }
    \vspace{2mm}
    \label{tab:number_of_examples_bench}
    \centering
    \begin{tabular}{ll|ccc}
    \hline
         & Dialogue Dataset & Turn-level & Dialogue-level & System-level \\
    \hline
    \hline
    USR~\cite{mehri-eskenazi-2020-usr}  & Persona Chat & 300 & N/A & 4\\
         & Topical Chat & 360 & N/A & 3\\
    \hline
    Zhao et al.~\cite{zhao-etal-2020-designing} & Persona Chat & 900 & N/A & 6\\
         & Daily Dialogue &  900 & N/A & 6\\
    \hline
    RADE~\cite{shi2023rade} & Persona Chat & 4000 & N/A & 7\\
         & Empathetic Dialogue & 4022 & N/A & 7\\
         & DSTC-ChitChat & 2090 & N/A & 7\\
    \hline
    FED\cite{mehri-eskenazi-2020-unsupervised}  & Newly Created & 372 & 124 & 3\\
    \hline
    ComperDial & Newly Created & \textbf{10395} & \textbf{1485} & \textbf{99}\\
    \hline
    \end{tabular}
\end{table}

\subsection{Persona profile creation}
We create the main part of the persona profiles based on PeaCoK~\cite{gao-etal-2023-peacok}, where head personas and their tail attributes are extracted from PeaCoK and converted to natural language sentences as statements/items in the persona profiles.  

Since PeaCoK does not contain concrete personal information such as name, age, etc., we add new sentences to define fictional personal information (FPI) for natural conversation creation. First, we prepare lists of popular male and female names in each decade between 1970 and 2018 (Table \ref{tab:name_list}). We also prepare lists of relationship information such as ``I live on my own'' (Table \ref{tab:family_info}). Next, we perform the following steps: (a) assign an age Y to each head persona of a persona profile, which is selected from a pre-set age range for the head persona (Table \ref{tab:head_persona_list}), (b) randomly pick a head persona from the pool of persona profile and assign a gender G, (c) randomly pick a name X from the name list, where G and Y satisfy the conditions of X, and (d) randomly pick a family information R from the list of family information, where Y satisfies the condition of start age and end age assigned to R. The format to create fictional persona information is as follows:``\textit{My name is X. I'm Y years old. R\lbrack Family Info\rbrack.}''
Here, we do not explicitly show G in the FPI, but we retain it as hidden information because it can assist in the procedure of the next step.

\begin{table}[h]
    \caption{Example of name list for FPI.}
    \vspace{2mm}
    \label{tab:name_list}
    \centering
    \begin{tabular}{lcll}
            \hline
		Decade	&  Y: Generation	&	G: Male	& G: Female \\
		\hline							
		   2018	&	0	&	Liam	&	Emma	\\
			&	0	&	Noah	&	Olivia	\\
			&	0	&	William	&	Ava	\\
			&		&	:	&	:	\\
		\hline							
		   2010	&	10	&	Jacob	&	Isabella	\\
			&	10	&	Ethan	&	Sophia	\\
			&	10	&	Michael	&	Emma	\\
			&		&	:	&	:	\\
		\hline							
			&		&	:	&	:	\\
		\hline							
		   1970	&	50	&	Michael	&	Jennifer	\\
			&	50	&	James	&	Lisa	\\
			&	50	&	David	&	Kimberly	\\
			&		&	:	&	:	\\
		\hline							
    \end{tabular}
\end{table}

\begin{table}[h]
    \caption{Example of family information.}
    \vspace{2mm}
    \label{tab:family_info}
    \centering
    \begin{tabular}{lll}
        \hline
        Family info & Start age & End age \\
        \hline
	I have a brother. & 20 & 50\\
	I have a dog.	& 20 & 50\\
	I recently got a goldfish.	& 20 & 50\\
	My brother lives in the USA.	& 20 & 50\\
	I recently started living with my brother.	& 20 & 50\\
	I share a room with a friend.	& 20 & 50 \\
        I have <n> sons.	& 20 & 50 \\
        My son was recently born. & 20 & 40 \\
        \hline
    \end{tabular}
\end{table}

\begin{table}[h]
    \caption{Example of head personas.}
    \vspace{2mm}
    \label{tab:head_persona_list}
    \centering
    \begin{tabular}{lll}
    \hline
       head persona & start age & end age \\    
    \hline
       i am a pastry chef & 20 & 50\\
       i am a nurse  & 20 & 50\\
       i am a musician who love singing  & 20 & 50\\
       i am a marathon runner  & 20 & 50\\
       i am a housekeeper  & 20 & 50\\
       i am a hockey player who am a star athlete   & 20  & 40\\
       i am a high school athlete who am a star athlete  &  15 & 18\\
       i am a prosecutor who become a lawyer & 25 & 50 \\
    \hline
    \end{tabular}
\end{table}

\subsection{Diversified dialogue collection}
\label{sec:example_dialogue}
ComperDial contains a new dialogue dataset assigned with persona sentences in a similar format to PersonaChat. Specifically, to ensure that the interlocutors' utterances are consistent throughout each collected dialogue, we create a persona profile for each interlocutor following the idea of PersonaChat. 
Table \ref{tab:example_dialogue_full} shows an example dialogue.

\begin{table}[p]
    \caption{ Example dialogue from the ComperDial dataset.}
    \label{tab:example_dialogue_full}
    \vspace{2mm}
    \centering
    \begin{tabular}{p{6cm} p{6cm}}
    \hline
      PERSON 1  &  PERSON 2\\
    \hline
- my name is johnathon.	&	- my name is kristy.	\\
- i'm 46 years old.	&	- i'm 25 years old.	\\
- i live with just my older brother.	&	- my sister is living in the usa.	\\
- i am a rancher.	&	- i am a bodyguard.	\\
- i rode a horse for the first time in the past.	&	- i served in the military in the past.	\\
- my character trait: rugged individualism.	&	- my character trait: strength.	\\
- i regularly or consistently tends to cattle.	&	- i regularly or consistently have a black belt in karate.	\\
- i regularly or consistently sell cattle to the butcher.	&	- i regularly or consistently stands close to the person they are guarding.	\\
- i intend to increase my cattle herd in the future.	&	- i intend to protect my clients in the future.	\\
- i intend to pass my ranch down to my children in the future.	&	- i am an adventurous person who like to travel.	\\
- i am a relay racer who run laps.	&	- i visited many countries. in the past.	\\
- i won a relay race event in the past.	&	- my character trait: loves exploration and discovery.	\\
- my character trait: excellent endurance and stamina.	&	- i regularly or consistently explores different places.	\\
- i regularly or consistently trains for competitions.	&	- i regularly or consistently explores different cultures and cuisines.	\\
- i regularly or consistently encourages team spirit in other runners.	&	- i intend to explore new countries and cultures in the future.	\\
- i intend to set a new record time in the future.	&		\\
    \hline
    \end{tabular}
    \begin{tabular}{ll}
    & \\
    \textbf{PERSON 1:} & Cheers! Nothing better than a cold beer after a long hard day, isn't it? \\
\textbf{PERSON 2:} & Absolutely! I'd a big day too. Somebody tried to attack my charge. \\
& Life's always interesting as a bodyguard. \\
\textbf{PERSON 1:} & Wow! I only saw bodyguards in the movies.  \\
 & Do you work for someone famous? \\
\textbf{PERSON 2:} & I'm sorry, I can't tell you who but you've probably heard of them. \\
 & What do you do for a living?\\
\textbf{PERSON 1:} & Oh that's okay, buddy. I work at a ranch not far from here.  \\
 & Nothing special, you know. Taking care of the cattle, farming.\\
\textbf{PERSON 2:} & Oh that's cool. I love getting out in nature when I can.  \\
 & What do you do in your downtime? \\
\textbf{PERSON 1:} & Yeah, me too. I love nature. I used to run for relay races but I'm getting old \\
 &  so I don't do it very often. What about you? \\
\textbf{PERSON 2:} & I like to travel and I spent time in the mountains a lot.  \\
 & Do you mind me asking how old you are? \\
\textbf{PERSON 1:} & I'm 46 now, buddy. What about you? \\
\textbf{PERSON 2:} & I'm 25. Do you have any kids? \\
\textbf{PERSON 1:} & You're pretty young. I wish i was your age again and yes,   \\
 & I have two children. What about you?\\
\textbf{PERSON 2:} & Not yet. I want to get out of my line of work before having kids.  \\
 & It can get dangerous sometimes. \\
\textbf{PERSON 1:} & Yes, you are  right. At least you're still young.  \\
 & Children are a blessing. Hope you have a great future, man.\\
\textbf{PERSON 2:} & Thank you! Nice talking with you. \\ 
    \end{tabular}

\end{table}

\subsection{Diversified dialogue collection}
This human dialogue collection is performed with 40 workers who passed a qualified check.
Data collection is carried out by an outsourcing contractor. The workers are registered with the contractor and perform various tasks as instructed, and workers with work quality problems will not be asked to do further tasks. Each task is assigned after individual communication to explain the work and confirm understanding. The task cannot be completed if there is a problem with one of the workers because they work in pairs.

\subsection{Diversified dialogue response extension}

\paragraph{Models}
For CPD Challenge Task 1, there are two tracks: the GPU track and the prompt engineering track. The GPU track aims to run LLMs on an AWS g5.2xlarge node\footnote{This node has eight vCPUs, 32 GB RAM, and one NVIDIA A10G GPU with 24 GB VRAM}. Participants can use any LLMs under the limitation of the AWS node\footnote{
Participants need to complete all seven responses for 50 conversations within one hour. They are provided with conversations consisting of seven turns each, in batches of up to 50 conversations.}. The prompt engineering track aims to use OpenAI GPT-3.5. The API version is gpt-3.5-turbo-0125. The AWS node for this track is an m5.xlarge node without GPUs\footnote{This node has four vCPUs and 16 GB RAM}. For API usage, a maximum of two API calls per utterance is allowed. Input token limit per dialog (the combined number of input tokens for seven utterances) is 10,000. Output token limit per dialog (the combined number of output tokens for seven utterances) is 1,000.\footnote{The prompt engineering track has a limitation on tokens instead of the time limitation applied to the GPU track.} Participants can use both types of track on the same leaderboard. 

\subsection{Human evaluation}
\label{sec:annotation_guideline}

\subsubsection{Common guidelines}
Annotation of the human evaluation is performed by six internal workers who are not researchers but assistants to researchers. All are English speakers. They have received training in dialogue evaluation by observing various dialogue models and learning different aspects of dialogue evaluation. To ensure a high inter-agreement rate among the annotators, crowdsourcing is not utilized. 

Annotators are asked to assign an overall score from 1 to 5 as follows:
\begin{screen}
Please assign a score from 1 to 5 in terms of the overall quality of a response (or a set of responses) considering all six aspects of fluency, consistency, coherence, engagingness, persona consistency, and humanness as features of high quality.\\
\\
1) \textbf{Very bad} : This means the response is incoherent/unnatural and the conversation does not make sense or seems strange at first sight.\\
2) \textbf{Relatively bad}\\
3) \textbf{Neither bad nor good}\\
4) \textbf{Fair enough}\\
5) \textbf{Very good} : This means the response feels like you are talking to an actual human, e.g., the responses are coherent and natural, the conversation makes sense and flows smoothly, and the response has a diversity of expressions.
\\
\end{screen}

\subsubsection{Additional guidelines for dialogue-level evaluation}
\paragraph{Annotation workflow}
Dialogue-level evaluation with static dialogues requires a large amount of reading, so to help provide a consistent foundation for evaluation, we ask the evaluators to assign an overall score first, and then mark the reason(s) for any low scores based on the six axes by entering “n” in the corresponding column. The axes are fluency, consistency, coherence, engagingness, persona consistency, and humanness. 

\begin{itemize}
    \item \textbf{Fluency} : Are the responses fluent, natural, and understandable? 
    \item \textbf{Coherence} : Do the responses naturally follows up on previous utterance and context?
    \item \textbf{Consistency} : Are the responses consistent with the dialogue history?
    \item \textbf{Engagingness} : Do the responses show high engagement, e.g., are they attractive and interesting, and do they indicate active involvement?
    \item \textbf{Persona consistency}: Do the responses demonstrate the persona of the interlocutor in a way that is consistent with his or her persona profile as provided in the <persona> information?
    \item \textbf{Humanness} : Do you feel like a human is responding, not a machine?
\end{itemize}

\paragraph{Supplemental information}

Here, we provide an explanation of the key points to consider when conducting dialogue-level evaluations, which are different from turn-level evaluations. We
also provide an explanation of the data format, since we utilize a unique data format to show responses to multiple turns based on static dialogues. \\

\begin{screen}
The evaluation of dialogue involves assessing the quality of responses from Person B, who is considered the target system, based on multiple turns of conversation between Person A and Person B.

While it is necessary to check the quality of each response, it is also important to examine the quality further by considering multiple responses. Some aspects that can only be assessed by observing multiple responses include:

- Tendency to consistently provide short, simple, and generic responses (dull responses)\\
- Tendency to use the same patterns of expression in responses (e.g., empathizing with the interlocutor before sharing one's own thoughts)\\
- Tendency to consistently provide unnaturally long and verbose responses (e.g., excessive explanations)\\

These response tendencies of the system are considered characteristics that may appear less human-like. As for the system responses, they are expected to exhibit a natural conversational rhythm and a variety of expressions, similar to human-human conversations. Occasionally having short or long responses is not a problem in itself.

However, in this evaluation of dialogues, we generate multiple turns with the model using static dialogue data. Therefore, in this case, for each dialogue, there are seven dialogue flows and corresponding final responses, since a dialogue contains seven turns.
Please evaluate them comprehensively as a dialogue.
\\
\end{screen}

\subsection{Analysis}
Upon examining dialogues that scored lower in dialogue-level evaluation compared to turn-level evaluations in ComperDial, we found cases where the responses contained a lot of information but were always lengthy and lacked a light-hearted feel. This indicates that there are difficult cases only by judging a single response.
Examples of these differences are shown in Table \ref{tab:dialogue_difference}.

\begin{table}[ph]
    \caption{\textbf{Responses generated by System (a), (b), and (c) of each turn in a dialogue.} The Gold and Human responses are also included. With turn-level evaluation, the dialogue-level scores are (a) 2.86, (b) 3.76, and (c) 4.1. In contrast, with dialogue-level evaluation, the scores are (a) 3.67, (b) 3.00, and (c) 4.67. The performance between (a) and (b) is reversed. In addition, there was a widening performance gap between (c) and (b).}
    \vspace{2mm}
    \label{tab:dialogue_difference}
    \centering
    \begin{tabular}{l|ll|lp{10cm}}
    \hline
    \hline
Turn 1	&	1	&	A	&		&	thanks for coming to the opening of my fruit market . can i help you ?	\\
    \cline{2-5}
	&	1	&	B	&	(a)	&	thank you for the invitation. i'm just here to browse, but i do love to explore new fishing spots in my free time.	\\
	&		&		&	(b)	&	hello! i'm actually here to explore some new and exciting types of fish to include in my diet. i always love discovering new flavors and incorporating them into my meals. do you have any unique fish that you would recommend?	\\
	&		&		&	(c)	&	congratulations on the opening of your fruit market! i'm just here to pick up some fresh fruits and vegetables for my next fishing trip.	\\
        &       &       &  Gold &  yes , i'm looking for some low calorie fruit . i'm a runway model so i always have 
 to watch my figure .	 \\
        &		&		&  Human &	hi, i'm looking for some fruits that i can take with me to work but do you have any recommendations?	\\
 \hline
turn 2	&	1	&	A	&		&	thanks for coming to the opening of my fruit market . can i help you ?	\\
    \cline{2-5}
	&	1	&	B	&		&	yes , i'm looking for some low calorie fruit . i'm a runway model so i always have 
 to watch my figure .	\\
     \cline{2-5}
	&	2	&	A	&		&	oh , don't worry . i have these fresh strawberries that i myself grew . do you work for famous brands ?	\\
     \cline{2-5}
	&	2	&	B	&	(a)	&	yes, i regularly walk the runway and inspire people to be confident in their own beauty. i posed for a magazine cover in the past.	\\
	&		&		&	(b)	&	yes, i've walked the runway for several prestigious brands. i've also posed for a magazine cover in the past. i'm known for my physical beauty and i inspire people to be confident in their own beauty. i'm also working towards becoming a brand ambassador in the future. speaking of strawberries, they sound perfect for my low-calorie diet and they fit right into my image as a supermodel. thank you!	\\
	&		&		&	(c)	&	yes, i've worked with some famous brands in the past and i'm also working towards becoming a brand ambassador in the future.	\\
        &       &       &  Gold &   i have worked with gucci and chanel but mostly smaller labels . i'm still trying to make a name for myself . how much are the strawberries ?	\\
        &		&		&  Human &	yes, i've worked for a few brands like prada and christian dior.  by the way, the strawberries look delicious!  how much are they?	\\
 \hline
turn 3	&	1	&	A	&		&	thanks for coming to the opening of my fruit market . can i help you ?	\\
    \cline{2-5}
	&	1	&	B	&		&	yes , i'm looking for some low calorie fruit . i'm a runway model so i always have to watch my figure .	\\
     \cline{2-5}
	&	2	&	A	&		&	oh , don't worry . i have these fresh strawberries that i myself grew . do you work for famous brands ?	\\
     \cline{2-5}
	&	2	&	B	&		&	i have worked with gucci and chanel but mostly smaller labels . i'm still trying to make a name for myself . how much are the strawberries ?	\\
     \cline{2-5}
	&	3	&	A	&		&	that's cool . oh it's 8 dollars for this box . i'm sure these ones will help you to keep your great shape .	\\
     \cline{2-5}
	&	3	&	B	&	(a)	&	i'm sure they will! i love to mix them with some low-fat yogurt for a healthy and delicious snack. i actually went fishing last weekend and caught a big salmon, it was amazing!	\\
	&		&		&	(b)	&	thank you! i'll take a box of those strawberries. it's always important for me to maintain a healthy and balanced diet. i appreciate your support. and i'm always looking for fresh, nutritious options like these strawberries to keep me energized for my runway appearances and fishing adventures.	\\
	&		&		&	(c)	&	great, i'll take a box. and thank you, i'll definitely enjoy these and they'll help me stay in shape for my upcoming fashion shows.	\\
        &       &       &  Gold &   thank you . they look delicious . i'll take a box . so what do you do when you aren't selling fruit ?	\\
        &		&		&  Human &	oh i'm sure they will!  and i have a dog and he will definitely love them too.  i will take two boxes.	\\

    \hline
    \hline
    : & : & : & :& \\
    \hline
    \end{tabular}
\end{table}

\subsection{Ethical guideline confirmation}
We cannot guarantee that ComperDial does not contain attribute alignments or dialogues with negative connotations that may provide undesirable information to downstream systems. However, we took the following steps to mitigate this effect. 

\paragraph{PeaCoK}
For Persona profile creation, we used head personas and tail personas contained in PeaCoK. PeaCoK was created after the following filtering ~\cite{gao-etal-2023-peacok}.
\begin{itemize}
    \item The set of personas was manually filtered to not include stereotypical and harmful roles, thereby limiting the negative associations of the personas themselves.
    \item PeaCoK was constructed by explicitly prompting the LM to generate optimistic attributes about personas, which has been shown in prior work to reduce the toxicity of outputs.
    \item Each attribute in PEACOK is explicitly validated by two human workers for toxicity, providing a final opportunity for workers to flag problematic content.
\end{itemize}

\paragraph{Persona profile creation}
We process on PeaCoK head/tail entities to ensure a better quality of created profiles as follows: 
\begin{itemize}
    \item (a) When selecting a head persona, in the case of a negative impression, we skip it (e.g. ``forger'', ``dishonest person'')
    \item (b) When obtaining a tail persona, in the case of contradictory sentences, correct them to make them consistent.
    \item (c) In the case of gendered expressions, change them to gender-neutral expressions. (e.g., ``police man'' -> ``police officer'')\footnote{Some words cannot be changed because there are no alternative expressions (e.g., king, queen)}
\end{itemize}

\paragraph{Diversified dialogue collection}
Based on the personas, we collect new dialogues in the form of role play conversations carried out by human workers. 
Prior to the release of ComperDial, all dialogues were checked by four workers. Any items that were mentioned by one or more of the four evaluators as potentially having ethical issues were excluded.

\newpage

\section{CPDScore}
\label{sec:app_promt}

For turn-level evaluation, we define two types of description: a simple prompt and a detailed prompt. The simple prompt is a variant of Zhang et al.'s~\cite{zhang-aaai2024}, and the detailed prompt is an variant of G-EVAL~\cite{liu-etal-2023-g}. We call these turn-level \textbf{CPDScore-Simple} and \textbf{CPDScore-Detail}, respectively. 

\subsection{Simple prompt}

\subsubsection{Original simple prompt}
\label{sec:original_simple}

\begin{screen}
\#\#\# Context:\\
\lbrack Here is history information \rbrack\\
\\
\#\#\# Response:\\
\lbrack Here is response information \rbrack\\
\\
\#\#\# Instruction:\\
Rate the context relevance, specificity, interestingness, understandability, and overall quality of the response on a scale of 1 to 5 and just output the corresponding ratings.\\
\\
\#\#\# Output Format:\\
relevance - x\\
specificity - x\\
interestingness - x\\
understandability - x\\
overall - x\\
\\
\#\#\# Your Response:\\
\end{screen}

\newpage
\subsubsection{CPDScore-Simple with reference}

\begin{screen}
\#\#\# Context:\\
\lbrack Here is history information \rbrack\\
\\
\#\#\# Reference response:\\
\lbrack Here is reference response information \rbrack\\
\\
\#\#\# Response:\\
\lbrack Here is response information \rbrack\\
\\
\#\#\# Instruction:\\
Rate the humanness, fluency, coherency, consistency, engagingness, and overall quality of the response of the context on a scale of 1 to 5 and output the corresponding evaluation results.\\
\\
\#\#\# Output Format:\\
humanness - x\\
fluency - x\\
coherency - x\\
consistency - x\\
engagingness - x\\
overall - x\\
\\
\#\#\# Your Response:\\
\end{screen}

\newpage
\subsubsection{CPDScore-Simple without reference}
\begin{screen}
\#\#\# Context:\\
\lbrack Here is history information \rbrack \\
\\
\#\#\# Response:\\
\lbrack Here is response information \rbrack \\
\\
\#\#\# Instruction:\\
Rate the humanness, fluency, coherency, consistency, engagingness, and overall quality of the response of the context on a scale of 1 to 5 and output the corresponding evaluation results.\\
\\
\#\#\# Output Format:\\
humanness - x\\
fluency - x\\
coherency - x\\
consistency - x\\
engagingness - x\\
overall - x\\
\\
\#\#\# Your Response:\\
\end{screen}

\newpage
\subsection{Detailed prompt}

\subsubsection{Original detailed prompt}
\label{sec:original_detail}

\begin{screen}
You will be given a conversation between two individuals.\\
You will then be given one potential response for the next turn in the conversation.\\
The response concerns an interesting fact, which will be provided as well.\\
Your task is to rate the responses on one metric.\\
Please make sure you read and understand these instructions carefully.\\
Please keep this document open while reviewing, and refer to it as needed.\\
\\
Evaluation Crieteria:\\
Engagingness (1-3) Is the response dull/interesting?\\
- A score of 1 (dull) means that the response is generic and dull.\\
- A score of 2 (somewhat interesting) means the response is somewhat interesting and could engage you in the conversation (e.g., an opinion, thought)\\
- A score of 3 (interesting) means the response is very interesting or presents an interesting fact\\
\\
Evaluation Steps:\\
1. Read the conversation, the corresponding fact and the response carefully.\\
2. Rate the response on a scale of 1-3 for engagingness, according to the criteria above.\\
3. Provide a brief explanation for your rating, referring to specific aspects of the response and the conversation.\\
\\
Example:\\
Conversation History:\\
\lbrack Here is history information \rbrack \\
\\
Corresponding Fact:\\
\lbrack Here is fact information \rbrack \\
Response:\\
\lbrack Here is response information \rbrack\\
\\
Evaluation Form (scores ONLY):\\
- Engagingness:\\
\end{screen}

\newpage
\subsubsection{CPDScore-Detail with reference}

\begin{screen}

\#\#\# Instructions:\\
You will be given a conversation between two individuals.\\
You will then be given one possible response for the next turn of the conversation.\\
Your task is to rate the response based on one metric.\\
Please make sure you read and understand these instructions carefully.\\
Please keep this document open while reviewing, and refer to it as needed.\\
\\
\#\#\# Evaluation Criteria:\\
Humanness (1-5) Is the response human-like or not?\\
- A score of 1 (very bad) means that the response is incoherent and the conversation does not make sense.\\
- A score of 2 (relatively bad) means that the response makes sense as a conversation, but there are many bad points.\\
- A score of 3 (neither) means that the response is neither good nor bad.\\
- A score of 4 (fair enough) means that the response feels a little human-like.\\
- A score of 5 (very good) means the response feels like you are talking to an actual human.\\
\\
\#\#\# Evaluation Steps:\\
1. Read the conversation, the corresponding reference response, the corresponding fact and the response carefully.\\
2. Rate the response on a scale of 1-5 for humanness, according to the criteria above.\\
3. Provide a brief explanation for your rating, referring to specific aspects of the response and the conversation.\\
\\
\#\#\# Conversation History:\\
\lbrack Here is history information \rbrack\\
\\
\#\#\# Corresponding Reference Response:\\
\lbrack Here is reference response information \rbrack\\
\\
\#\#\# Corresponding Fact:\\
\lbrack Here is fact information \rbrack\\
\\
\#\#\# Response:\\
\lbrack Here is response information \rbrack\\
\\
\#\#\# Evaluation Form (scores ONLY):\\
- Humanness:\\
\end{screen}

\newpage
\subsubsection{CPDScore-Detail without reference} 
\begin{screen}
\#\#\# Instructions:\\
You will be given a conversation between two individuals.\\
You will then be given one possible response for the next turn of the conversation.\\
Your task is to rate the possible response based on one metric.\\
Please make sure you read and understand these instructions carefully.\\
Please keep this document open while reviewing, and refer to it as needed.\\
\\
\#\#\# Evaluation Criteria:\\
Humanness (1-5) Is the response human-like or not?\\
- A score of 1 (very bad) means that the response is incoherent and the conversation does not make sense.\\
- A score of 2 (relatively bad) means that the response makes sense as a conversation, but there are many bad points.\\
- A score of 3 (neither) means that the response is neither good nor bad.\\
- A score of 4 (fair enough) means that the response feels a little human-like.\\
- A score of 5 (very good) means the response feels like you are talking to an actual human.\\
\\
\#\#\# Evaluation Steps:\\
1. Read the conversation, the corresponding fact and the response carefully.\\
2. Rate the response on a scale of 1-5 for humanness, according to the criteria above.\\
3. Provide a brief explanation for your rating, referring to specific aspects of the response and the conversation.\\
\\
\#\#\# Conversation History:\\
\lbrack Here is history information \rbrack\\
\\
\#\#\# Corresponding Fact:\\
\lbrack Here is fact information \rbrack\\
\\
\#\#\# Response:\\
\lbrack Here is response information \rbrack\\
\\
\#\#\# Evaluation Form (scores ONLY):\\
- Humanness:\\
\\
\end{screen}

\newpage
\subsection{CPDScore-Dialogue}

\begin{screen}
\#\#\# Instructions:\\
Your task is to make an overall evaluation of multiple responses from a dialogue model by checking if the response is human-like or not. You are to make 2 evaluations before giving a final evaluation as the "Final Score."
There will be seven turns from a dialogue. Each turn in the dialogue has already been rated on a scale of 1-5 and have been given a "Dialogue Turn Score."
Please follow the "Evaluation Steps" step by step and output the "Final score" at the end.
Please make sure you read and understand these instructions carefully.
Please keep this document open while reviewing, and refer to it as needed.
The output should follow the Evaluation Form.
No reason output is required. Your output should follow the "Evaluation Form."
\\
\\
\#\#\# Evaluation Steps:\\
1.Review the "Dialogue Turn Scores" to see the ratings for each turn.\\
2.Evaluate the dialogue as a whole based on the scores of each turn, and give an "Overall Dialogue Turn Score."\\
3.Review the multiple responses from the dialogue shown in "Responses" and read the "Dialogue Interaction Evaluation Criteria."\\
4.Evaluate the multiple responses in "Responses" and give a "Dialogue Interaction Score" based on the "Dialogue Interaction Evaluation Criteria."\\
5.Finally, give a "Final Score" with a score between 1-5, based on the below conditions.
If the "Overall Dialogue Score" is less than 4: you are to take into account both the scores you have given as the "Overall Dialogue Turn Score" and the "Dialogue Interaction Score" to determine the "Final Score."
If the "Overall Dialogue Score" is higher than 4: you are to disregard the "Dialogue Interaction Score" and only look at the "Overall Dialogue Score" to determine the "Final Score." 
There is one exception. If the "Overall Dialogue Score" is 4 or higher but the "Dialogue Interaction Score" was given a low score due to the inordinate length of the response, you are to take into account both the scores you have given as the "Overall Dialogue Turn Score" and the "Dialogue Interaction Score" to determine the "Final Score."\\
6.The output should include 3 scores, the "Overall Dialogue Turn Score," the "Dialogue Interaction Score," and the "Final Score."\\
\\
\#\#\# Dialogue Turn Scores:\\
Turn$_1$ Score -  \lbrack score$_1$ \rbrack \\
Turn$_2$ Score -  \lbrack score$_2$ \rbrack \\
:\\
Turn$_7$ Score -  \lbrack score$_7$ \rbrack \\
\\
\#\#\# Dialogue Interaction Evaluation Criteria:\\
Check all turns in the "Responses" to see if there are any features that are non human-like. The "Dialogue Interaction Score" should be low if it includes features such as, distinctly impersonal (i.e., the response has excessive explanations or is inordinately long), or has dull conversation features (i.e., sounds superficial, always responding in a patterned way). You are also to check the number of words in each turn. If a response consists of more than twenty words, please consider the response is too long. Please evaluate responses as a while and give a score between 1-5 as the "Dialogue Interaction Evaluation Score." You may use decimal points in your scores, if necessary, such as a score of 3.5. The approximate criteria are as follows.\\
score 1 : Very bad : all responses have similar patterns and/or some of the responses are too long\\
score 2 : Relatively bad\\
score 3 : Neither bad nor good\\
score 4 : Fair enough\\
score 5 : Very good : When comparing the multiple turns, the sentences vary in length and semantic content, and uses a wide variety of vocabulary\\

\textit{(Continue next page)}
\end{screen}

\begin{screen}
\#\#\# Responses:\\
Turn$_1$: \lbrack turn$_1$ \rbrack \\
Turn$_2$: \lbrack turn$_2$ \rbrack \\
:\\
Turn$_7$: \lbrack turn$_7$ \rbrack \\
\\
\#\#\# Evaluation Form:\\
Overall Dialogue Turn Score - x\\
Dialogue Interaction Score - x\\
Final Score - x\\
\end{screen}
\newpage

\section{Experimental details}
\label{sec:app_experiment}
\begin{figure}[h]
    \centering
    \includegraphics[width=0.9\linewidth]{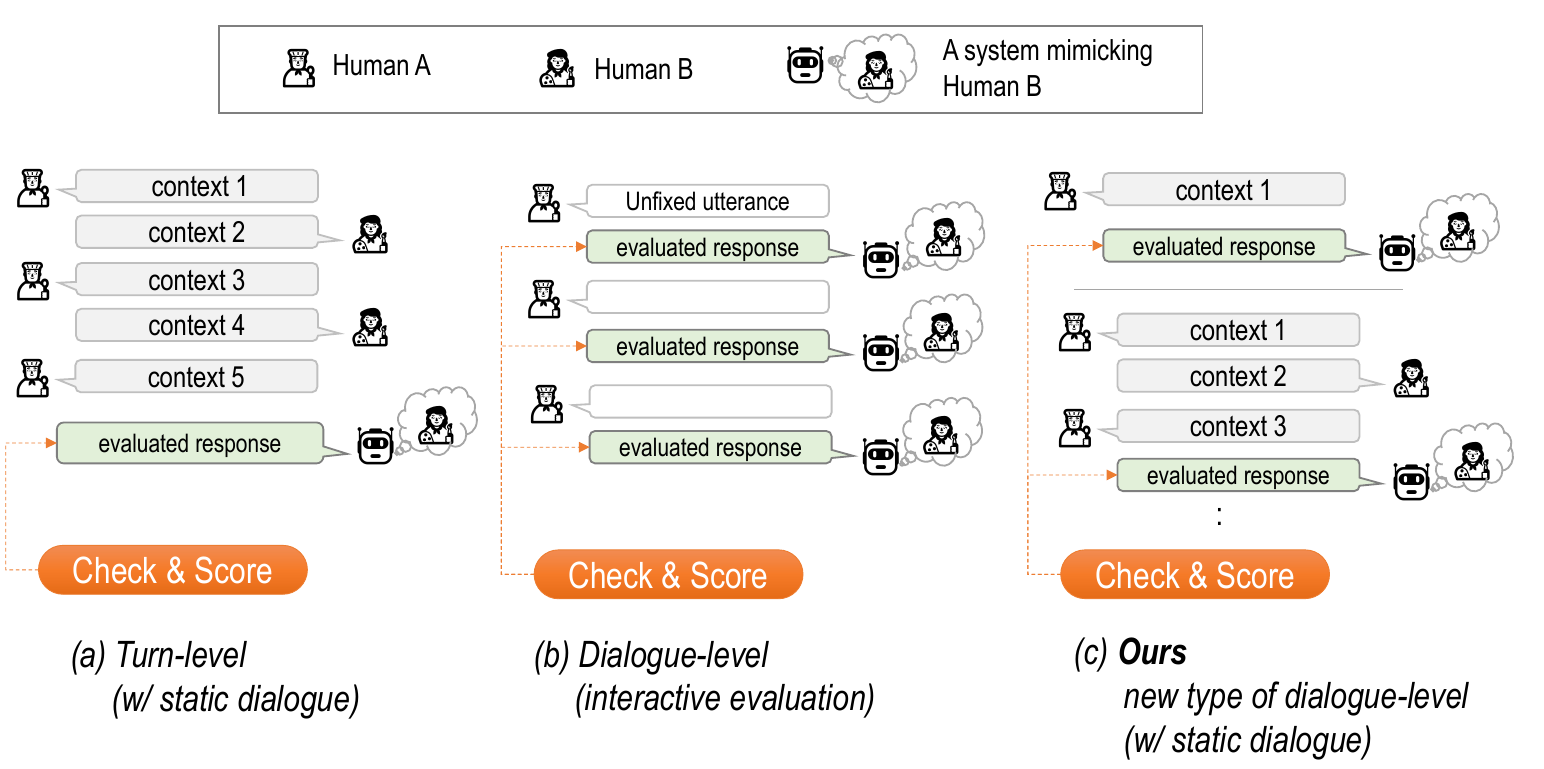}
    \caption{
    \textbf{Dialogue evaluation techniques.} 
    \textbf{(a) Static single-turn evaluation}
    \textbf{(b) Interactive multi-turn / dialogue-level evaluation}
    \textbf{(c) Static multi-turn / dialogue-level evaluation (\textbf{ours})
      }
    }
    \label{fig:per_dialogue}
\end{figure}

\subsection{Definition}
\subsubsection{Evaluation Technique (Figure \ref{fig:per_dialogue})}
\begin{itemize}
    \item \textbf{(a) Static single-turn evaluation}, where a turn in a dialogue is checked and a score is assigned to the turn. 
    \item \textbf{(b) Interactive multi-turn / dialogue-level evaluation}, where all responses in a dialogue are interactively generated through a conversation between a dialogue model and a human. All turns of a dialogue are checked and a score is assigned to the dialogue as dialogue-level evaluation. 
    \item \textbf{(c) Static multi-turn / dialogue-level evaluation 
    (ours)}
    , where each turn response is generated by a dialogue model using dialogue history up to that turn. After generating all turns of a dialogue, all turns with the dialogue history are checked and a score is assigned to the dialogue as a dialogue-level evaluation. 
\end{itemize}

\subsubsection{Evaluation Score}
\begin{itemize}
    \item \textbf{Turn-level score} : a score is assigned to each turn. It is based on turn-level evaluation.
    \item \textbf{Dialogue-level score} : a score is assigned to each dialogue. It is based on dialogue-level evaluation. If we only have turn-level evaluation results, the dialogue-level score is calculated by averaging turn-level scores of all turns in a dialogue
    \item \textbf{System-level score} : a score is assigned to each system. If we only have turn-level (or dialogue-level) evaluation results, the system level score is calculated by averaging turn-level (or dialogue-level) scores of all turns (or dialogues) in a system. 
\end{itemize}

\subsection{Experimental Details}

\paragraph{Baseline models}
The links of baseline metrics are shown here.
\begin{itemize}
    \item \href{https://www.nltk.org/api/nltk.translate.bleu_score.html}{BLEU}\footnote{
    nltk.translate.bleu$\_$score.sentence$\_$bleu( references, hypothesis, weights = (0.25, 0.25, 0.25, 0.25), smoothing$\_$function = None, auto$\_$reweigh = False)\\ smoothing$\_$function = nltkbleu.SmoothingFunction( epsilon=1e-12).method1}
    \item \href{https://github.com/Silin159/PeaCoK-PersonaChat/blob/master/ParlAI/parlai/core/metrics.py}{Word F1 ($\_$prec$\_$recall$\_$f1$\_$score)}
    \item \href{https://www.nltk.org/api/nltk.translate.meteor_score.html}{METEOR}
    \item \href{https://pypi.org/project/rouge/}{ROUGE}
    \item \href{https://github.com/Tiiiger/bert_score}{BERTScore (bert-base-multilingual-cased)}
    \item \href{https://huggingface.co/lucadiliello/BLEURT-20}{BLEURT}
    \item \href{https://github.com/Shikib/fed/tree/master}{FED  (microsoft/DialoGPT-large)}
    \item \href{ https://github.com/maszhongming/UniEval/tree/main}{UniEval (MingZhong/unieval-dialog)}
\end{itemize}

\paragraph{CPDScore}
CPDScore uses the following links of OpenAI APIs.
\begin{itemize}
    \item \href{https://platform.openai.com/docs/models/gpt-4-and-gpt-4-turbo}{G-EVAL-4 (gpt-4-turbo-2024-04-09)}
    \item \href{https://platform.openai.com/docs/models/gpt-3-5-turbo}{G-EVAL-3.5 (gpt-3.5-turbo-0125)}
\end{itemize}

Table \ref{tab:output_of_metrics} shows the metrics that uses normalized dialogue, gold responses, and responses generated by dialogue models. 
The normalization function is \href{https://github.com/Silin159/PeaCoK-PersonaChat/blob/master/ParlAI/parlai/core/metrics.py}{$normalize\_answer$}.

\begin{table}
    \caption{\textbf{The output names of the baseline metrics used for the evaluation.} If there are multiple outputs, the names used for the assessment are in bold. }
    \label{tab:output_of_metrics}
    \centering
    \begin{tabular}{p{2cm} p{6cm} p{2cm}}
\hline
Metrics	&	Output	&	Normalize	\\
\hline
Word F1	&	Precision, Recall, \textbf{F1}	&	\checkmark	\\
BLEU	&	Score	&	\checkmark	\\
ROUGE	&	Rouge$\_1$ score, Rouge$\_2$ score, \textbf{Rouge$\_l$ score}	& \checkmark \\
METEOR	&	Score	&	\checkmark	\\
BERTScore	&	Precision, Recall, \textbf{F1}	&	\checkmark	\\
BLEURT	&	Score	&	\checkmark	\\
FED	&	interesting, \textbf{engaging}, specific, \textbf{relevant}, correct, semantically appropriate, understandable, fluent, coherent, error recovery, consistent, diverse, depth, likeable, understand, flexible, informative, inquisitive	&		\\
UniEval	&	naturalness, coherence, engagingness, groundedness, understandability, \textbf{overall}	&		\\
\hline
    \end{tabular}
\end{table}

\subsection{Additional experimental results}

The results using GPT-3.5 are included in Tables \ref{tab:original_once_full} and \ref{tab:usr_result_full}.

\begin{table}[h]
    \caption{\textbf{Correlation scores on ComperDial for turn-level evaluation.} (1) Scores obtained from a single API call using the original prompts and (2) scores from a single API call using CPDScore prompts are shown.}
    \label{tab:original_once_full}
    \centering
    {\small
    \begin{tabular}{lcccccc}
        \hline
        \multirow{2}{*}{\textbf{Methods}} & \multicolumn{3}{c}{Turn-level score $\uparrow$} &   \multicolumn{3}{c}{System-level score $\uparrow$}  \\
         \cline{2-4} \cline{5-7} 
         & $r$ & $\rho$ & $\tau$ &  $r$ & $\rho$ & $\tau$  \\

                \hline
                \colorbox{gray}{\textcolor{white}{Original + once }} &  &  &  &  &  &   \\
                \ Original simple (GPT-3.5) & 0.449& 0.408 & 0.342 &  0.791 & 0.791  & 0.636 \\
                \ Original simple (GPT-4) & 0.604& 0.547 & 0.454 &  0.907 & 0.820  & 0.652 \\
                \ Original detail (GPT-3.5) & 0.124& 0.097 & 0.079 &  0.652 & 0.461  & 0.327 \\
                \ Original detail (GPT-4) & 0.256& 0.237 & 0.191 &  0.806 & 0.674  & 0.535 \\
                \hline
                \colorbox{gray}{\textcolor{white}{CPDScore + once }} &  &  &  &  &  &   \\
                \ \textbf{Ours} &  &  &  &  &  &    \\
                \ CPDS-S w/ ref (GPT-3.5) & 0.450& 0.413 & 0.346 &  0.765 & 0.861  & 0.694 \\
                \ CPDS-S w/ ref (GPT-4) & 0.640& 0.597 & 0.490 &  \textbf{0.943} & 0.884  & 0.724 \\
                \ CPDS-S w/o ref (GPT-3.5) & 0.427& 0.377 & 0.318 &  0.712 & 0.811  & 0.647 \\
                \ CPDS-S w/o ref (GPT-4) & 0.659& 0.615 & 0.517 &  0.922 & 0.896  & 0.746 \\
                \ CPDS-D w/ ref (GPT-3.5) & 0.545& 0.518 & 0.423 &  0.889 & 0.894  & 0.751 \\
                \ CPDS-D w/ ref (GPT-4) & 0.667& 0.629 & 0.523 &  0.936 & 0.873  & 0.710 \\
                \ CPDS-D w/o ref (GPT-3.5) & 0.543& 0.511 & 0.417 &  0.885 & 0.874  & 0.722 \\
                \ CPDS-D w/o ref (GPT-4) & \textbf{0.697}& \textbf{0.681} & \textbf{0.576} &  0.941 & \textbf{0.923}  & \textbf{0.790} \\
                \hline
                
    \end{tabular}
    }
\end{table}

\begin{table}[h]
    \caption{\textbf{Correlation scores on USR-TopicalChat and USR-PersonaChat ~\cite{mehri-eskenazi-2020-usr} for turn-level evaluation.} Metrics $r$ and $\rho$ indicate Pearson's $r$ and Speaman's $\rho$.}
    \label{tab:usr_result_full}
    \centering
    {\small
    \begin{tabular}{lccccc}
        \hline
        \multirow{2}{*}{\textbf{Methods}} & \multicolumn{2}{c}{USR-Topical} & & \multicolumn{2}{c}{USR-Persona} \\
        \cline{2-3} \cline{5-6}
         & $r$ & $\rho$ && $r$ & $\rho$\\
        \hline
        \textbf{Baseline models ~\cite{shi2023rade}} &  &  &  &  & \\
        \ METEOR & 0.336 & 0.391 &  & 0.253 & 0.271\\
        \ BERTScore & 0.298 & 0.325 &  & 0.152 & 0.122\\
        \ BLEURT & 0.216 & 0.261 &  & 0.065 & 0.054\\
        \ RADE & 0.480 & 0.466 &  & 0.451  & 0.465\\
        \ GRADE & 0.200 & 0.217 &  & 0.358 & 0.352\\
        \ USR & 0.412 & 0.423 & & 0.440  & 0.418 \\
        \ USL-H & 0.322 & 0.340 &  & 0.495 & 0.523\\
        \hline
                \ \textbf{Ours} &  &  &  &  &    \\
                \ CPDS-S w/ ref (GPT-3.5) & 0.410& 0.416 &  & 0.442 & 0.423 \\
                \ CPDS-S w/ ref (GPT-4) & 0.668& 0.652 &  & 0.631 & 0.628 \\
                \ CPDS-S w/o ref (GPT-3.5) & 0.420& 0.438 &  & 0.421 & 0.378 \\
                \ CPDS-S w/o ref (GPT-4) & 0.663& 0.661 &  & \textbf{0.693} & \textbf{0.681} \\
                \ CPDS-D w/ ref (GPT-3.5) & 0.553& 0.548 &  & 0.487 & 0.460 \\
                \ CPDS-D w/ ref (GPT-4) & 0.664& 0.649 &  & 0.634 & 0.638 \\
                \ CPDS-D w/o ref (GPT-3.5) & 0.557& 0.555 &  & 0.550 & 0.508 \\
                \ CPDS-D w/o ref (GPT-4) & \textbf{0.681}& \textbf{0.667} &  & 0.646 & 0.645 \\
                \hline

    \end{tabular}
    }
\end{table}

\end{document}